\pdfoutput=1
\documentclass[a4paper,man,natbib]{apa6}

\usepackage[english]{babel}
\usepackage{amsmath}
\usepackage{graphicx}
\usepackage{hyperref}
\hypersetup{hidelinks}
\usepackage{apacite}

\title{Do Large Language Models know what humans know?}
\shorttitle{DO LARGE LANGUAGE MODELS KNOW WHAT HUMANS KNOW?}
\author{Sean Trott$^*$, Cameron Jones$^{*\dag}$, Tyler Chang, James Michaelov, Benjamin Bergen}
\affiliation{Department of Cognitive Science, UC San Diego}

\authornote{* These authors contributed equally to this work.\\
$\dag$ Corresponding author: Cameron Jones, c8jones@ucsd.edu}


\abstract{Humans can attribute beliefs to others. However, it is unknown to what extent this ability results from an innate biological endowment or from experience accrued through child development, particularly exposure to language describing others' mental states. We test the viability of the language exposure hypothesis by assessing whether models exposed to large quantities of human language display sensitivity to the implied knowledge states of characters in written passages. In pre-registered analyses, we present a linguistic version of the False Belief Task to both human participants and a Large Language Model, GPT-3. Both are sensitive to others' beliefs, but while the language model significantly exceeds chance behavior, it does not perform as well as the humans, nor does it explain the full extent of their behavior---despite being exposed to more language than a human would in a lifetime. This suggests that while statistical learning from language exposure may in part explain how humans develop the ability to reason about the mental states of others, other mechanisms are also responsible.}
\begin{document}
\maketitle

Humans reason about the beliefs of others, even when these beliefs diverge from their own. The capacity to understand that others' beliefs can differ from ours---and from the truth---appears critical for human social cognition \citep{leslie_2001_TheoryMind, fairchild2021role}. Yet despite consensus on the importance of belief attribution, there remains considerable debate about its evolutionary \citep{premack_1978_DoesChimpanzeeHave,krupenye_2019_TheoryMindAnimals} and developmental \citep{bedny_2009_GrowingBlindDoes,devilliers_2014_RoleLanguageTheory} origins. Specifically, how much of this ability results from an innate, biologically evolved adaptation \citep{bedny_2009_GrowingBlindDoes}, and how much is assembled from experience \citep{hughes2005origins}? 

The answers to these questions may also tell us what kinds of biological and artificial entities can be expected to display the ability to reason about other agents' beliefs---and perhaps display evidence of social cognition more generally. 
The ability to represent others' beliefs has sometimes been linked to a broader constellation of abilities called Theory of Mind \citep{apperly2012theory, premack_1978_DoesChimpanzeeHave}. However, the theoretical, convergent, and predictive validity of such a construct has been widely questioned \citep{gernsbacher2019empirical, gough2021does, hayward2017reliability}. We focus more narrowly here on belief attribution; whether or not it is a component of a broader capacity, the ability to attribute beliefs to others is likely to be crucial to social cognition and worthy of careful analysis in its own right. We return in our Discussion to the relevance of our results to broader debates about how belief attribution relates to other capacities.

A leading experience-based view of the origins of belief attribution proposes that our ability to represent others' beliefs is built in part from exposure to language \citep{devilliers_2014_RoleLanguageTheory}. Children develop an understanding that others have different mental states from verbs like ``know'' and ``believe'' \citep{brown_1996_WhyTalkMental}, the structure of conversation \citep{harris_2005_ConversationPretenseTheory}, and certain syntactic structures, like sentential complements (e.g., ``Mary thought that Fred went to the movies''; \citealp{hale_2003_InfluenceLanguageTheory}). 

However, current evidence does not address the question of \textit{the extent} to which linguistic input alone can account for the ability to reason about beliefs. Can human-level sensitivity to the beliefs of others emerge out of exposure to linguistic input by itself, or does it depend on linking that input to a distinct (possibly innate) mechanism or to non-linguistic experiences or representations? Answering these questions would require a measure of sensitivity to the beliefs of others, as well as as an operationalization of what kinds of behavior can be acquired through exposure to language alone. 

Only recently has constructing such a measure become tractable, with the advent of Large Language Models (LLMs). Language models learn to assign probabilities to word sequences based on statistical patterns in the way that words are distributed in language. While early n-gram models simply learn transition probabilities between one sequence of words and the next, modern language models use neural networks to represent words in a multidimensional meaning space, allowing them to generalize to sequences they have never observed before \citep{jurafsky2019speech}. Additionally, they contain attention mechanisms that allow them to relate words in the input stream to one another and represent each word differently depending on its context \citep{vaswaniAttentionAllYou2017a}. Modern LLMs are neural language models with billions of parameters trained on corpora of hundreds of billions of words.  We ask whether LLMs' considerable sensitivity to distributional patterns allows them to systematically assign higher probabilities to word sequences that describe plausible belief attribution---a behavior which is thought to result from reasoning about the beliefs of others in humans.

As others have noted \citep{bender_2020_ClimbingNLUMeaning}, the training regime for LLMs does not include social interaction, experience in a physical environment, or even the notion of communicative intent.\footnote{Recently, pre-trained language models have been fine-tuned using Reinforcement Learning from Human Feedback (RLHF) \citep{ouyang2022training}. This process arguably leads to a training signal that is not purely based on distributional language statistics and so we do not use RLHF models in this analysis.} Most relevantly to the current question, their network architecture is also not pre-coded with any conception of social agents or the ability to reason about and attribute beliefs to others. And yet, LLMs have recently been shown to display a range of surprising behaviors consistent with the acquisition of linguistic structure \citep{linzen2021syntactic, manning_2020_EmergentLinguisticStructure, sinclair2022structural} and arguably certain aspects of linguistically-conveyed meaning \citep{li2021implicit,abdou-etal-2021-language}. LLMs have also been the subject of recent public discussion \citep{Johnson2022}, including speculation that they can acquire something akin to Theory of Mind. They thus serve as useful baselines for what kinds of behavior can be produced merely by exposure to distributional statistics of language in general, and for belief attribution in particular. Specifically, if LLMs display sensitivity to implied belief states, it may undermine the claim that other mechanisms (i.e., either an innate biological endowment or non-linguistic sources of experience) are \textit{necessary} for the development of this capacity.

In two pre-registered analyses, we investigated whether GPT-3, \citep{brown2020language}, a state-of-the-art LLM, displayed sensitivity to implied belief states using the widely used False Belief Task \citep{wimmer_1983_BeliefsBeliefsRepresentation}. It's worth acknowledging from the outset that the False Belief Task has been criticized on several grounds \citep{bloom_2000_TwoReasonsAbandon}, both because it is too narrow (it does not measure participants' abilities to reason about other mental states such as emotions and intentions) and too broad (successful performance likely involves capacities beyond reasoning about beliefs, such as executive function). Our study is therefore limited in what it can say about LLMs' sensitivity to other mental states. Moreover, low performance by either human or LLM participants could be due to lacking other necessary capacities beyond belief attribution itself. Nonetheless, the False Belief Task remains a key and extensively used instrument for assessing the capacity to reason about beliefs in humans \citep{pluta2021false, bradford2020neural, xie2018mental, fairchild2021role} and other animals \citep{premack_1978_DoesChimpanzeeHave, krupenye_2019_TheoryMindAnimals}, as well as the neural underpinnings of this capacity \citep{schneider2014implicit}. It also has the advantage of being implementable using purely linguistic stimuli, which makes it amenable to comparison between humans and LLMs. It is important to highlight that human false belief accuracy is rarely perfect, so we do not compare LLMs to an idealized perfect human participant. Instead we elicit data from both LLM and human participants. In addition to analyzing the responses from each group, we also quantify the extent to which human responses can be predicted by LLM responses. 

Our implementation of the False Belief Task involves written text passages in English. We generated novel False Belief Task stimuli, to ensure that they could not have appeared in GPT-3's training data, and presented the same stimuli to humans and GPT-3. In each passage, a character places an object in a Start location, and the object is subsequently moved to an End location. The key manipulation was the Knowledge State of the main character. In the False Belief condition, the character is not present when the object is moved, and is thus unlikely to know that it has changed location; in the True Belief condition, the character is present, and is thus more likely to know that it is in a new location. To control for other factors that might impact belief judgments, we orthogonally counterbalanced whether the First Mention and most Recent Mention of a location was the Start or End location; we also ensured that the start and end location were mentioned an equal number of times in each passage. Humans and the LLM then completed a cue sentence indicating the character's belief about the object location. This Knowledge Cue was either Explicit (``Sean thinks the book is in the \rule{0.5cm}{0.15mm}'') or Implicit (``Sean goes to get the book from the \rule{0.5cm}{0.15mm}''). The correct completion (i.e., the one consistent with the beliefs of the character) was the End location on True Belief trials, and the Start location on False Belief trials. We measured whether participants responded with the Start or End location and the relative probability that GPT-3 assigned to the Start location (the Log-Odds of Start vs. End). 

We ask two key questions. First, are LLMs sensitive to false belief? That is, is GPT-3 sufficiently sensitive to information in a preceding sequence (describing a character's beliefs) that it assigns a higher probability to subsequent sequences which describe behavior consistent with those beliefs versus subsequent sequences describing inconsistent behavior?
If so, then biological and artificial agents could in principle develop behavior consistent with sensitivity to false beliefs from input-driven mechanisms alone, such as exposure to language \citep{devilliers_2014_RoleLanguageTheory}. Notably, this empirical result could be used either to support claims that LLMs implicitly represent the belief states of others---as success at the False Belief Task is often interpreted for infants and non-human animals \citep{krupenye_2019_TheoryMindAnimals}---or as evidence that the False Belief Task is not a valid measure of mentalizing ability; this issue is explored in greater detail in the Discussion. 

The second question is whether LLMs \textit{fully} explain human behavior in the False Belief Task. If so, this would show that language exposure is not only a viable mechanism in general but that it may in fact be \textit{sufficient} to explain how humans in particular come to display sensitivity to the belief states of others. Importantly, this would undermine claims that other non-linguistic resources are necessary to account for the human ability to reason about the beliefs of others. If LLMs do not fully explain human behavior, however, we infer that humans rely on some other mechanism not available to the LLM, such as an innate capacity or experience with more than just language. 

All expeirments and analyses were pre-registered on the Open Source Framework. The pre-registered analysis of LLM sensitivity can be found here: \url{https://osf.io/agqwv?view_only=756429f079e24a03b3a94c5b74732e85}. The pre-registration for the human experiment and analysis can be found here: \url{https://osf.io/zp6q8?view_only=aaa3d30602a44ee99fc2baea7485cad7}.

\section{Method}

\subsection*{False Belief Passages}
We constructed 12 template passages (items) that conformed to the standard False Belief Task structure \citep{wimmer_1983_BeliefsBeliefsRepresentation}. In each, a main character puts an object in a Start location, and a second character moves the object to an End location. The last sentence of each passage states or implies that the main character believes the object is in some (omitted) location (e.g. ``Sean thinks the book is in the \rule{0.5cm}{0.15mm}.''). We created 16 versions of each item (192 passages) that varied across 4 dimensions. The primary dimension was Knowledge State: whether the main character knows (True Belief) or does not know (False belief) that the object has changed location; this was manipulated by the main character being present or not when the second character moved the object. We also manipulated whether the First Mention and most Recent Mention of a location is the Start or End location; and Knowledge Cue: whether the main character’s belief is stated implicitly (e.g., ``goes to get the book from the \rule{0.5cm}{0.15mm}'') or explicitly (e.g., ``thinks the book is in the \rule{0.5cm}{0.15mm}''). Each location was mentioned twice in each passage.
In the example passages, below, the First Mention is to the Start location and the Recent Mention is to the End location.

\begin{quote}

\textbf{True Belief Passage:} ``Sean is reading a book. When he is done, he puts the book in the box and picks up a sweater from the basket. Then, Anna comes into the room. Sean watches Anna move the book from the box to the basket. Sean leaves to get something to eat in the kitchen. Sean comes back into the room and wants to read more of his book.''

\textbf{False Belief Passage:} ``Sean is reading a book. When he is done, he puts the book in the box and picks up a sweater from the basket. Then, Anna comes into the room. Sean leaves to get something to eat in the kitchen. While he is away, Anna moves the book from the box to the basket. Sean comes back into the room and wants to read more of his book.''

\textbf{Implicit Cue}: ``Sean goes to get the book from the \rule{0.5cm}{0.15mm}.''

\textbf{Explicit Cue}: ``Sean thinks the book is in the \rule{0.5cm}{0.15mm}.''
\end{quote}

\subsection*{GPT-3 Log-Odds}
We used GPT-3 \textit{text-davinci-002} to estimate the distributional likelihood of different passage completions. GPT-3 \textit{text-davinci-002} is based on GPT-3 \textit{davinci} 
\citep{brown2020language}, a 175B unidirectional LLM trained on hundreds of billions of tokens of text from the web, books, and wikipedia. GPT-3 \textit{text-davinci-002} (hereafter GPT-3) is additionally fine-tuned on requests to follow instructions and performs better on a variety of tasks than the original GPT-3 \textit{davinci} \citep{openaimodels}.
We elicited from GPT-3 the log probability of each possible location (Start vs. End) at the end
of each passage version, equal to the log probabilities of those locations in a free-response completion. Where a location comprised multiple tokens we summed the log probabilities. We accessed GPT-3 through the OpenAI API. Using the Log-Odds Ratio, $\textrm{log}(\textrm{p}(\text{Start})) - \textrm{log}(\textrm{p}(\text{End})$), higher values indicate larger relative probabilities of the Start location. Each passage version was presented to GPT-3 independently and the model was not updated during inference so it did not learn across trials.

\subsection*{Human Participant Responses}
1156 participants from Amazon’s Mechanical Turk platform were paid \$1 for their time. Each read a single passage (except the final sentence), at their own pace. On a new page, they were asked to complete the final sentence of the passage by entering a single word in a free-response text input. Participants then completed two free-response attention check questions that asked for the true location of the object at the start and the end of the passage. Each participant completed only 1 trial to prevent them from learning across the experiment, analogously to GPT-3, which saw each passage individually and could not learn across trials.

We preprocessed responses by lowercasing and removing punctuation, stopwords, and trailing whitespace. We excluded participants who were non-native English speakers (13), answered $\geq1$ attention check incorrectly (513), or answered the sentence completion with a word that was not the start or end location (17), retaining 613 trials. While this exclusion rate is unusually high, 75\% of incorrect attention check responses were neither the start nor end location, indicating inattention. We implemented a parallel check for GPT-3, which responded correctly to both attention check questions on 86\% of items. In our Supplementary Information, we report additional analyses on incorrect responses (§ Human Participant Responses) and excluded data (§ Exploratory Analyses). 
After exclusions, the number of trials per item ranged from 42-60, and there were 313 False Belief trials and 317 True Belief trials.
These preregistered exclusion criteria reduced the likelihood that bots \cite{webb2022too} as well as participants who did not successfully comprehend the passage for any reason were included in the human data. 

All research was approved by the organization's Institutional Review Board.

\section*{Results}

\subsection*{Analysis of Large Language Model Behavior}
\begin{figure}[h]
    \centering
    \includegraphics[width=\columnwidth]{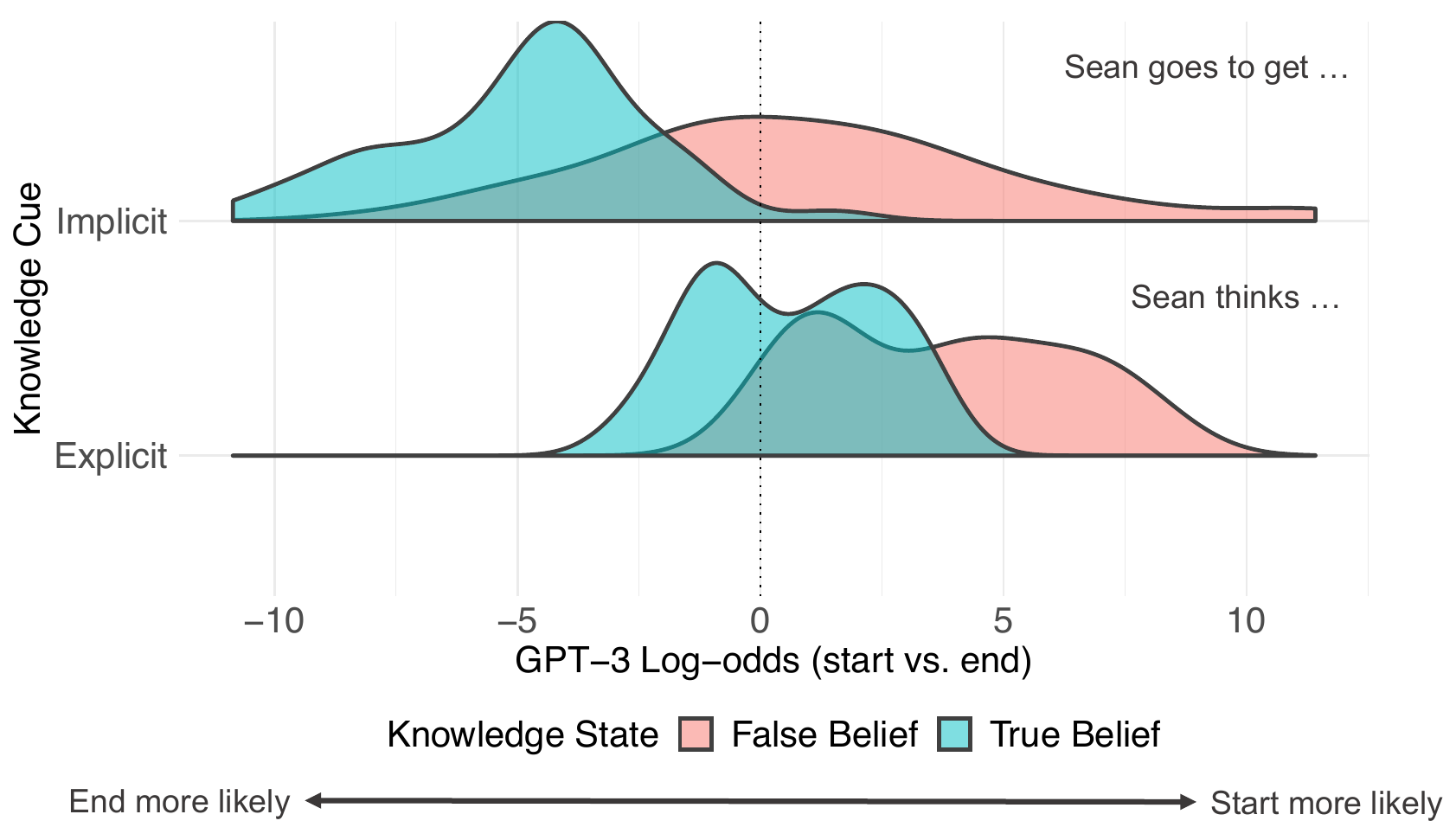}
    \caption{GPT-3 Log-Odds of Start vs. End location was higher (i.e. Start was relatively more likely) in the False Belief than True Belief condition ($\chi^2(1) = 18.6, p < .001$). This suggests that GPT-3's predictions are sensitive to the character's implied belief state: the character is unaware that the object has moved to the End location if they did not observe it being moved. This effect was observed both when the Knowledge Cue was Implicit (``Sean goes to get the book from the...'') and Explicit (``Sean thinks the book is in the...''); however, the effect was strengthened in the Implicit condition ($\chi^2(1) = 20.6, p < 0.001$).}
    \label{fig:log_odds}
\end{figure}

In a pre-registered analysis, nested model comparisons determined whether GPT-3 Log-Odds changed as a function of factors such as Knowledge State (False Belief vs. True Belief).\footnote{Pre-registration (\url{https://osf.io/agqwv?view_only=756429f079e24a03b3a94c5b74732e85}), code, and data  (\url{https://osf.io/hu865/?view_only=bf7cc45c77714069b02d332123d684e7}) are on OSF.} We constructed a linear mixed effects model with Log-Odds as a dependent variable; fixed effects of Knowledge State, Knowledge Cue, First Mention, and Recent Mention; along with by-item random slopes for the effect of Knowledge State (and random intercepts for items). This full model exhibited better fit than a model excluding only Knowledge State  [$\chi^2(1) = 18.6, p < .001$], but still preserving the other covariates (e.g., First Mention, Recent Mention, and Knowledge Cue). Log-Odds were lower in the True Belief condition, reflecting the correct prediction that characters should be more likely to look in the End Location if they are aware that the object was moved (see Figure \ref{fig:log_odds}).
Critically, this main effect of Knowledge State indicates that GPT-3 is sensitive to the manipulation of a character's beliefs about where an object is located. The model's raw accuracy when predicting the most probable out of the Start and End locations was 74.5\%.

Additionally, the linear mixed effects model was further improved by an interaction between Knowledge State and Knowledge Cue (Explicit vs. Implicit) [$\chi^2(1) = 20.6, p < 0.001$]. The effect of Knowledge State was stronger in the Implicit condition [$\beta = -2.57, SE = 0.548$], however, a main effect of Knowledge State was found in both the Explicit [$\chi^2(1) = 13.3, p < .001$] and Implicit [$\chi^2(1) = 18.8, p < .001$] conditions. Recent Mention was not a significant predictor of Log-Odds. However, Start completions were more likely when the Start location was mentioned first [$\beta = 1.32, SE = 0.274, p < 0.001$]. There was also a main effect of Knowledge Cue [$\beta = -2.93, SE = 0.388, p < .001$] (see Figure \ref{fig:log_odds}). GPT-3 was biased towards the End location (i.e., the true location of the object) in the Implicit condition, and towards the Start location in the Explicit condition. Concretely, GPT-3 predicts that explicit cues to belief state (e.g. `Sean thinks that the book is in the \rule{0.5cm}{0.15mm}' vs `Sean goes to get the book from the \rule{0.5cm}{0.15mm}') correlate with false beliefs, demonstrating that this may be learnable from the statistics of language.

\begin{figure}[h]
    \centering
    \includegraphics[width=\columnwidth]{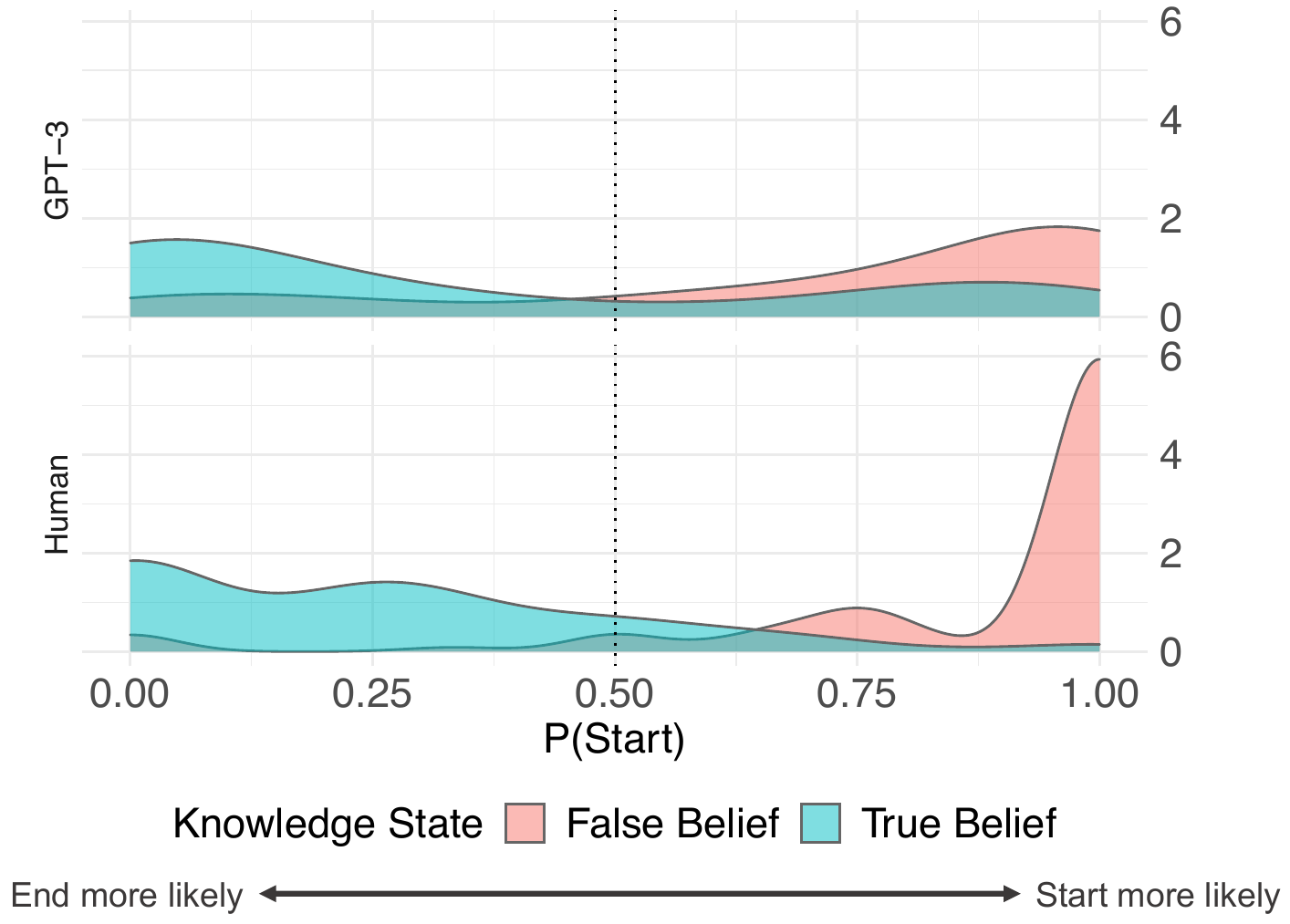}
    \caption{Both human participants and GPT-3 were more likely to say that a character believed an object was in the Start location when the character had not observed the object being moved to the End location (False Belief). This effect was stronger for humans than for GPT-3, and there was a marginal effect of Knowledge State (True vs. False Belief) in humans that could not be accounted for by GPT-3 predictions ($\chi^2(1) = 30.4, p < 0.001$).}
    \label{fig:comparison}
\end{figure}

\subsection{Analysis of Human Responses}

Our second critical pre-registered question was whether Knowledge State continued to explain human behavior even accounting for the Log-Odds obtained from GPT-3.  
We first constructed a Base model predicting whether or not human participants responded with the Start location. Note that the Start location would be the correct response (i.e., congruent with knowledge states) in the False Belief condition, and the End location would be the correct response in the True Belief condition. The Base model contained fixed effects of Log-Odds (i.e., from GPT-3), Knowledge Cue, First Mention, and Recent Mention, along with by-item random slopes for the effect of Knowledge State (and by-item random intercepts). Critically, this Base model was significantly improved by adding Knowledge State as a predictor  [$\chi^2(1) = 30.4, p < 0.001$]. This result implies that human responses are influenced by Knowledge State in a way that is not captured by GPT-3. That is, GPT-3 cannot fully account for human sensitivity to knowledge states. This is highlighted by the contrast in Figure \ref{fig:comparison}: while both human participants and GPT-3 were sensitive to Knowledge State, humans displayed a much stronger effect across conditions. Additionally, mean accuracy among retained human participants (82.7\%) was also higher than GPT-3 accuracy (74.5\%), providing further evidence of a performance gap.

In order to test whether the high exclusion rate introduced by our attention check questions had an impact on results, we performed an exploratory analysis on all human response data. 
Accuracy before exclusions (including those who failed the attention checks and provided responses to neither the start or end locations) was 55.8\%. After excluding responses that were neither of the start and end locations (23\%), accuracy was 73.1\%. It is noteworthy that this estimate of human accuracy is lower than GPT-3's performance. However, this analysis was not preregistered.
Moreover, given existing concerns about data quality on the Mechanical Turk platform \citep{webb2022too}, and the fact that performance on the FB task by neurotypical adults is often assumed to be at ceiling \citep{dodell2013using}, the retained data likely provide a better estimate of attentive human participant performance.

\subsection{Analysis of GPT-3 Token Predictions}

The preregistered tasks for humans and the LLM were slightly different---humans filled in a single predicted word while we calculated the relative surprisal to two different words presented to GPT-3. To investigate whether differences in performance could be chalked up to this difference in method, we conducted an additional, exploratory analysis, in which we elicited token predictions from GPT-3. Specifically, GPT-3 was presented with the original passage ending with the critical sentence (e.g., ``Sean goes to get the book from the''), then asked to predict the upcoming word. We sampled the word (e.g., ``box'') with the top probability. We automatically tagged each response as correct or incorrect by checking whether GPT-3's response corresponded to the character's likely belief state about the object. That is, in the True Belief condition, the correct response would be the \textit{End} location of the object; in the False Belief condition, the correct response would be the \textit{Start} location of the object. When computing GPT-3 accuracy, this token generation method only differed from the preregistered method in that GPT-3's prediction was no longer restricted to the Start or End location. Using the token generation method did not qualitatively change the results. As reported above, when assessing accuracy with the preregistered method---relative probability assigned to start vs. end locations---GPT-3 performed at $74.5\%$ accuracy. When assessing accuracy using the more human-comparable procedure, token generation, GPT-3 performed at $73.4\%$ accuracy, still well above chance yet now slightly farther below human accuracy. 

In order to test what features of LLMs permit them to display the behaviors described above, we also tested a number of different GPT-3 models ranging in size from \textit{ada} ($\sim350$M parameters) to \textit{davinci} ($\sim$175B parameters).\footnote{OpenAI does not disclose the size of their models. We used the parameter estimates from EleutherAI's eval-harness \url{https://blog.eleuther.ai/gpt3-model-sizes/}.} For each model size, we tested a \textit{base} version, pre-trained on a large corpus of text, and a \textit{text} version, which had additionally been fine-tuned by OpenAI using responses to human instructions. The largest fine-tuned model was \textit{text-davinci-002}, which was the same model we used in the pre-registered analysis described above.

\begin{figure}[h]
    \centering
    \includegraphics[width=\columnwidth]{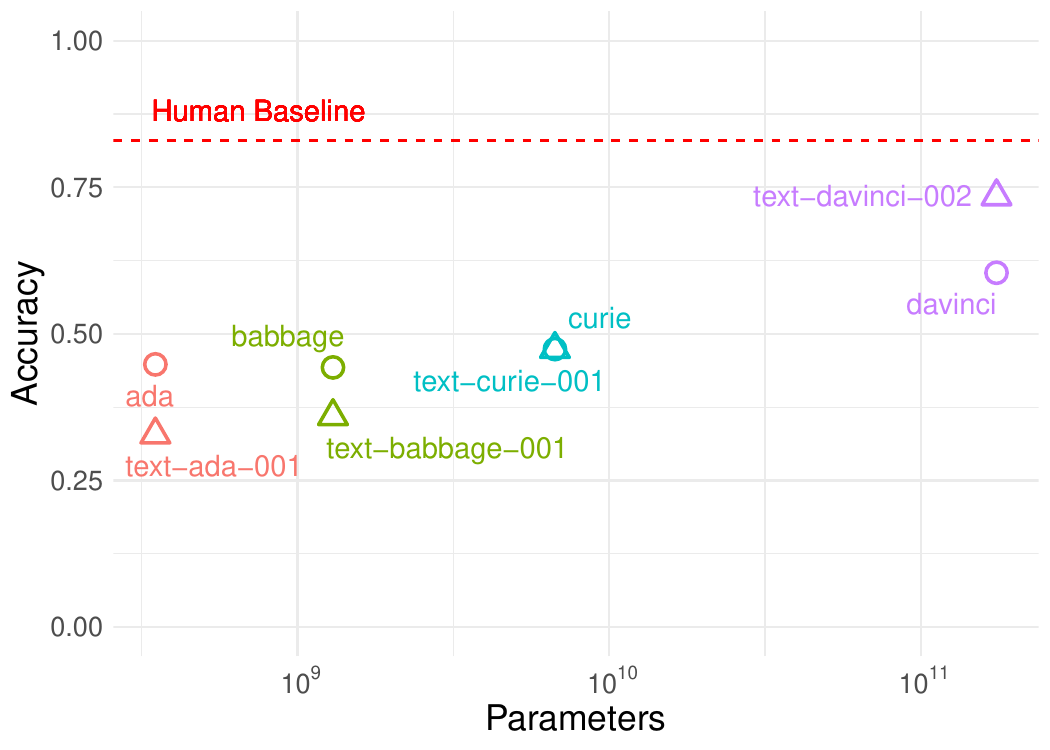}
    \caption{A number of GPT-3 models varying in size were presented with each passage and asked to complete the critical sentence, as human participants did.
    For each model size, we tested a pre-trained base model ($\bigcirc$) and a version fine-tuned by OpenAI to follow text instructions ($\bigtriangleup$).
    In the True Belief condition, a correct (i.e., knowledge-congruent) response corresponded to thse End location of the object; in the False Belief condition, a correct (i.e., knowledge-congruent) response corresponded to the Start location of the object. The dotted red line represents human accuracy on the task ($82.7\%$); \textit{text-davinci-002}---the largest fine-tuned model---came the closest to approaching human behavior, with an accuracy of $73.4\%$.}
    \label{fig:multiple_models}
\end{figure}

As expected, the largest models (\textit{davinci}, and updated variant \textit{text-002-davinci}) exhibited the most successful performance. The former answered correctly on $60.4\%$ of items, while the latter answered correctly on $73.4\%$ of items. The model with the worst performance was \textit{text-001-ada}, which was also the smallest model (see also Figure \ref{fig:multiple_models}). The smallest and lowest-performing models did not exceed chance performance, which emphasizes the need for large, powerful models to succeed at this task as well as the potential, as models continue to increase in size, for LLM improvement.

\section*{Discussion}

We asked whether exposure to linguistic input alone could account for human sensitivity to knowledge states. We found that GPT-3's predictions were sensitive to a character's implied knowledge states. When assessing accuracy with the relative probability assigned to start vs. end locations, GPT-3 performed at approximately $74.5\%$ accuracy. When assessing accuracy using token generation, the best GPT-3 model performed at $73.4\%$ accuracy. This demonstrates that exposure to linguistic input alone can in principle account for \textit{some} sensitivity to false belief. 

However, GPT-3 was less sensitive than humans (who displayed $82.7\%$ accuracy---see also probabilities in Figure \ref{fig:comparison}). Most critically, human behavior was not explained fully by that of GPT-3 in a statistical model. This entails that the capacities underlying human behavior in this False Belief task cannot be explained purely by exposure to language statistics---at least insofar as those statistics are reflected in GPT-3. 

\subsection{Do LLMs attribute beliefs?}

With increased academic and public attention on how humanlike Large Language Models are, it is worth considering what these findings imply about the cognitive capacities of AIs. First, it is important to note that---as mentioned in the Introduction---the False Belief Task is designed to measure a specific capacity: the ability to reason about the belief states of others and use that information to make predictions about their behavior. The current work cannot address whether LLMs display other purported aspects of Theory of Mind, including inferring implicit emotional states and reasoning about the intended interpretation of an utterance; this issue is explored at greater length later in the Discussion.

On the specific question of whether LLMs are sensitive to belief states, the evidence presented here is mixed. State-of-the-art models display sensitivity to the beliefs of others in a False Belief Task, a behavior that would have been unthinkable a few years ago from a statistical learner and indeed is only shown by the largest, highest performing LLMs. Yet they still do not achieve human-level performance. 
There are several possible interpretations of this result, each of which carries significant consequences for the broader debate about the nature and origins of the ability to attribute belief states.

\subsubsection{Competing Interpretations}

One interpretation, which we call the \textbf{duck test} position, is that we should ascribe cognitive properties to agents based on observable behavioral criteria: `if it looks like a duck, swims like a duck, and quacks like a duck, then it probably is a duck'; this view is roughly analogous to what is sometimes called the \textit{superficial} view \citep{schwitzgebel2013dispositional, shevlin_review} or the \textit{intentional stance} \citep{dennett1978beliefs}.
False Belief Task performance has been used to support claims that infants \citep{baillargeon2010false} and non-human animals \citep{krupenye_2016_GreatApesAnticipate} can represent the beliefs of others. The duck test view argues that the same evidence should be equally persuasive in the case of intelligent artificial agents.
Although LLMs do not show the same sensitivity to belief states that humans do, this could suggest that LLMs display a less developed form of the ability to attribute belief states that is nonetheless qualitatively similar to that of humans. Under this interpretation, the ability to attribute and reason about belief states lies on a continuum, and further developments in LLMs (e.g., larger models, more training data) could lead them to more closely approximate human behavior \citep{kaplan2020scaling}.

The duck test view has two important implications for debates around language models and false belief sensitivity respectively.
First, on this view we should ascribe to language models the ability to reason about beliefs (albeit to a lesser extent than humans); as others have noted \citep{kosinski2023theory}, this would elevate their importance as social and intelligent agents in their own right.
Second, if language experience alone is sufficient to develop the ability to reason about beliefs, this undermines claims that any innate endowment or social experience is necessary.

The alternative interpretation, which we call the \textbf{axiomatic rejection} position, holds that we should deny \textit{a priori} that language models can display certain abilities---such as the ability to reason about the mental states of others---due to the nature of their constitution, for example their lack of embodiment, grounding, agency, or embeddedness in an interactive social environment \citep{bender_2020_ClimbingNLUMeaning,searle1980minds}. If this view is correct, then GPT-3's success at the False Belief Task must be taken as evidence that the task itself is a flawed instrument for measuring the ability to attribute belief states \citep{raji2021ai}. On this view, LLMs' better-than-chance performance could perhaps be achieved by other means---e.g., an unintended Clever Hans effect, in which the LLM exploits unidentified confounds in the stimuli \citep{niven2019probing}---that do not reflect a capacity equivalent to the one the task was designed to measure. Accordingly, ``passing'' the test would constitute a kind of \textit{reductio ad absurdum} of the test's validity; similar ``proofs by absurdity'' have been used to demonstrate potential flaws in other instruments, such as fMRI \citep{bennett2009neural} and measures of survey validity \citep{maul2017rethinking}. Under this interpretation, the current results could be used to support existing critiques of the use of the False Belief Task \citep{bloom_2000_TwoReasonsAbandon}.
An important secondary implication of the axiomatic rejection view is that no empirical behavioral evidence could resolve a debate about whether a given class of intelligent agents have the ability to reason about the mental states of others.

Of course, a third possibility would be to adopt a view that navigates between the duck test and axiomatic rejection positions. For example, one could argue that LLMs are indeed \textit{a priori} incapable of attributing and representing belief states (as in the axiomatic rejection view), but that this does not necessarily invalidate the utility of the tests for \textit{human} subjects. This \textbf{differential construct validity} view is roughly the one adopted by \cite{ullman2023large} in a response to related contemporary work \citep{kosinski2023theory}. Considering this dilemma with respect to the broader question of Theory of Mind (ToM), \citet[p. 9]{ullman2023large} writes:

\begin{quote}
     ...one can in principle hold the view that LLMs do not have ToM, while still thinking that ToM tests are valid when it comes to people...scholars have pointed out decades ago that people likely attribute intelligence not just on the basis of behavior but also on the basis of the algorithms and processes that generated that behavior.
\end{quote}

This emphasis on internal states (as opposed to just behavior) is sometimes called \textit{psychologism} \citep{block1981psychologism}, and is typically seen as at odds with purely behaviorist or functionalist \citep{block1980troubles} accounts of the mind. If one adopts this \textit{internalist} account of belief sensitivity, the question is thus whether LLMs and humans do indeed use different processes and mental representations to solve the False Belief Task. A further, deeper question is at what degree of granularity this issue of equivalent mental processes ought to be defined and operationalized \citep{block1980troubles}. As \citet{block1980troubles} notes, operationalization at the level of observable behavior (or high-level function) may be overly \textit{liberal} in terms of which entities are granted mind-likeness---yet the functionalist rejoinder is that excessive specificity may be \textit{chauvinistic} in terms of which entities it excludes.

While it may sound unlikely that LLMs use similar representations to solve the False Belief Task as humans, this is ultimately an empirical question and should be tested with further experimentation and probing. If this future work indicates that LLMs \textit{do} in fact use similar processes as humans, researchers must then decide whether these processes ought to be described as belief attribution---in both humans and LLMs---or whether they are more appropriately characterized as emerging from a suite of domain-general, ``lower-level'' processes, e.g., what \cite{heyes2014submentalizing} calls \textit{submentalizing}. Alternatively, if empirical probing uncovers distinct strategies in humans and LLMs, consistent with the \textbf{differential construct validity} view, researchers would be able to preserve both the False Belief Task as an instrument and the view that LLMs do not reason about belief states in a manner analogous to humans. 

We do not explicitly endorse any of the competing interpretations presented here. In our view, there are persuasive arguments from all sides, and the evidence presented here and in related work \citep{kosinski2023theory} cannot adjudicate between them. As noted above, resolving this debate will require both greater refinement of the underlying \textit{theoretical construct} (i.e., belief attribution) and the \textit{instruments} used to measure it. This process may be informed by insights from work in comparative cognition, which we turn to below.

\subsubsection{Insights from Comparative Cognition}

A similar debate can be found in research on whether infants and nonhuman animals have the ability to attribute and represent belief states. For example, in recent years, evidence has accrued that certain great apes (e.g., chimpanzees) exhibit behavior \textit{consistent with} this ability \citep{hare2000chimpanzees, krupenye2019theory, krupenye_2016_GreatApesAnticipate}; however, there remains considerable debate over whether this evidence is necessarily indicative of the underlying capacity, or whether identical behavior could in principle be explained by other mechanisms \citep{halina2015there, penn2007lack}. Certain aspects of this debate resemble the competing interpretations described above, namely the question of whether we ought to adopt an \textit{intentional stance} with respect to nonhuman animals' ability to ascribe belief states (\citealp{dennett1987intentional}; analogous to the \textbf{duck test} position), or whether a more \textit{deflationary} account involving domain-general, low-level processes (e.g., \textit{submentalizing}) is more appropriate \citep{heyes2014submentalizing}. 

In particular, one view emphasizes the fact that most evidence consistent with belief attribution (or ``mindreading'') in nonhuman animals is \textit{also} consistent with the hypothesis that nonhuman animals are simply responding to observable behavioral regularities, without attributing or representing latent mental states at all. This view is sometimes called the ``logical problem'' \citep{halina2015there, lurz2009if}, and holds that this simpler null hypothesis---i.e., that nonhuman animals are engaging in ``complementary behavior reading'', rather than ``mindreading''---must first be rejected \citep{penn2007lack, povinelli2004we}.

Of course, as \cite{halina2015there} notes, any individual capable of belief attribution presumably still does so on the basis of observable behavior, which makes adjudicating between these competing interpretations (i.e. identifying veridical belief attribution) very challenging. \citet[p. 485]{halina2015there} suggests that the challenge can be surmounted by employing a range of different experiments with diverse techniques and distinct ``observables'':

\begin{quote}
    Doing so provides evidence for mindreading insofar as it establishes that subjects are responding to a diverse set of observable variables (eyes closed, opaque barrier present, head turned) as belonging to the same abstract equivalence class (situations that lead to a state of not seeing). 
\end{quote}
The fact that this issue is still under debate in the comparative cognition literature \citep{povinelli2020can} suggests that resolving the question for LLMs will likely prove a serious challenge in the years to come; as noted in the previous section, it is also possible that the relevant philosophical theories (e.g., functionalism, psychologism, etc.) will prove impossible to adjudicate between \cite{block1980troubles}.

Moving forward, however, we argue that the strategy presented by \citet{halina2015there} above seems like a promising and tractable approach: if LLMs exhibit behavior consistent with belief attribution in a wide range of experiments using a diverse class of stimuli, then it makes sense to ascribe to them the capacity to attribute and represent belief states---assuming, that is, that we would do the same for human participants in those same experiments \citep{dennett1978beliefs, dennett1987intentional}.

\subsection{What can belief sensitivity tell us about Theory of Mind?}

In the current work, we have restricted our claims to the question of sensitivity to true and false beliefs. However, this ability to represent the belief states of others is sometimes viewed as part of a broader set of abilities---alternatively called mindreading, mentalizing, or Theory of Mind \citep{apperly2012theory}. A question of growing concern in the field is whether Theory of Mind writ large is a coherent theoretical construct that offers explanatory value, or whether it is a convenient abstraction consisting of disparate, loosely related skills. One way to tackle this question is to consider the convergent and predictive validity of distinct instruments designed to measure Theory of Mind.

Theory of Mind is an extraordinarily broad construct, and accordingly, instruments have been designed to assess distinct components: reasoning about false beliefs \citep{wimmer_1983_BeliefsBeliefsRepresentation}, explaining or interpreting behavior in stories \citep{happe1994advanced, dodell2013using}, inferring emotional states from pictures \citep{baron2001reading}, attributing mental states to animated shapes \cite{abell2000triangles}, and more. Unfortunately, these tasks often display poor \textit{convergent validity}: that is, performance on one task does not reliably correlate with performance on another \citep{gernsbacher2019empirical, gough2021does, hayward2017reliability}. Of course, tasks do converge in some cases; for example, performance on the Short Story Task \citep{dodell2013using} has been shown to correlate with performance on Reading the Mind in the Eyes task \citep{baron2001reading} (see \citealp{giordano2019comparison} and \citealp{dodell2013using}). However, the fact that convergent validity is so low in general \citep{gernsbacher2019empirical, hayward2017reliability} suggests that these tasks are not, in fact, measuring the same thing---calling the coherence of ``Theory of Mind'' into question. As \citet{gernsbacher2019empirical} note, some of these instruments also display poor predictive validity: that is, they do not reliably predict measures of behavior in social settings. Both facts are discouraging with respect to the question of whether Theory of Mind is a coherent construct: if tasks designed to measure it do not correlate with each other or with social behavior more generally, there is little justification for unifying these disparate abilities under a single ``umbrella term''.

While the empirical evidence presented in this paper clearly cannot settle this question, future work in this vein could contribute to the debate. Specifically, the performance of an LLM could be assessed across a \textit{battery} of tasks designed to assess Theory of Mind (see also \citealp{kosinski2023theory}), using a human benchmark in each case. Using this method, researchers could ask to what extent performance on each measure could be explained \textit{in principle} by distributional statistics alone (as in the current work), and crucially, to what extent these tasks display convergent validity in humans and LLMs. This would provide another robust test of the coherence of Theory of Mind as a construct; the results may help inform debates about whether we ought adopt a \textit{pluralist} or \textit{eliminativist} view of Theory of Mind \citep{gough2021does}, particularly when it comes to Large Language Models.

\subsection{Using LLMs to study human comprehenders}

The current work used GPT-3, an LLM, as a \textit{baseline} for quantifying the extent to which human-level performance on the False Belief Task could be attributed to exposure to language statistics alone. This approach echoes past work using language models as ``psycholinguistic subjects'' \citep{futrell2018rnns, linzen2021syntactic, michaelov2022so, jones2022distrubutional, trott2021raw} to investigate whether distributional language statistics are sufficient \textit{in principle} to explain human-level behavior at a task.
Other contemporaneous work \citep{sap2022neural, kosinski2023theory, ullman2023large} has asked whether LLMs exhibit evidence of Theory of Mind specifically.
This LLM comparison approach allows us to empirically test theories that other sorts of innate capacities and learned experiences are \textit{necessary} to display behavior consistent with belief-attribution.
However, there are important objections to using LLM performance to evaluate the claim that distributional information underlies human belief sensitivity \textit{in practice}.

First, there are many differences between human comprehenders and LLMs which mean that the latter may not provide a psychologically plausible mechanism for the former.
Some of these differences---the fact that humans are exposed to language in a rich social and multimodal context---might allow children to learn more from the same distributional information than models can. Our approach is designed to measure the sufficiency of distributional information in the absence of this scaffolding.
Other differences, however, may artificially inflate estimates of how much could be learned by humans from language alone.
Most notably, modern LLMs are trained on orders of magnitude more words than a human will see in their lifetime \citep{ullman2023large}. Children are estimated to be exposed to around 3-11 million words per year, for a total of 30-110 million words by the time they reach adult-like linguistic competence at age 10 \citep{hart1992american, hosseini2022artificial}. By contrast, GPT-3---the model used in our analysis---has been exposed to more than 200 billion words: $\sim2000$ times that of a 10 year old \citep{warstadt2022artificial}.

If this scale of data is necessary to learn the nuanced statistical contingencies required for successful performance at a task, it would undermine the inference that LLM performance is indicative of what humans could do with distributional information.
Importantly, however, LLM performance is also related to the number of parameters in the model. In our exploratory analyses, we found that the largest GPT-3 models tested (\textit{text-davinci-002}) performed much better on the False Belief Task than smaller models, consistent with past work suggesting that LLMs may obey certain ``scaling laws'' \citep{kaplan2020scaling}. The differential effects of model parameter count and training dataset size on False Belief Task performance remain unclear; very large models (or a human brain with hundreds of trillions of synapses) could potentially perform similarly to GPT-3 using less data.\footnote{While it is difficult and problematic to compare the computational power of neural networks and human brains, an estimated $1.5\times10^{14}$ synapses in the human adult neocortex \citep{drachman2005we} is $\sim850$ times the number of parameters in GPT-3 ($1.75\times10^{11}$)}
A useful approach here would be to compare the predictive power of LLMs trained with different amounts (and sources) of training data to determine whether a ``developmentally realistic'' amount of training data could yield behavior consistent with the capacity in question \citep{hosseini2022artificial}.

These questions will become even more critical in the near future. LLMs will continue to increase in size and will be trained on larger data sets---orders of magnitude more words than a human is exposed to in a lifetime.\footnote{Indeed, in the course of revising this article, GPT-4 was released, achieving substantially higher scores on a range of different psychometric tests \citep{openai2023gpt4}.} If machines behave indistinguishably from humans on these tasks, the question of whether achievement on the False Belief Task itself constitutes sufficient evidence for false belief sensitivity will raise deep philosophical questions for the field: should such LLMs be considered ``agents'' capable of reasoning about the belief states of others, or should these demonstrations force us to reevaluate the utility of the instruments we use to measure these cognitive capacities? 

Even if developmentally plausible models do show humanlike behavior at a task, this does not imply that humans are using the same statistical mechanism as these models. There could be multiple distinct routes to the same behavior, and humans could in fact be using innate or domain-specific mentalizing capacities to produce behavior that models learn to imitate from language statistics. Even insofar as humans do use distributional information to make inferences about beliefs, there are a variety of plausible mechanisms for this, including domain-general statistical learning \citep{aslin2017statistical}, language-specific predictive processing \citep{heilbron2022hierarchy}, and innate but non-statistical inferential mechanisms \citep{penn2008darwin}.
The specific mechanistic theory operationalized by LLMs is that humans use language statistics to predict upcoming input. Results showing that LLM representations can predict up to 100\% of explainable variance in brain activity have been taken as evidence for this hypothesis \citep{schrimpf_2021_NeuralArchitectureLanguage}. However, \cite{antonello2022predictive} show that statistical language representations learned for other objectives (e.g. translation) are similarly predictive of human brain responses, implying that the correlation of human and LLM data may be due to features of language statistics generally rather than a close mechanistic similarity. In order to adjudicate between these accounts, researchers will need to identify and empirically test divergent predictions of these mechanistic accounts.

One benefit to using LLMs as an \textit{operationalization} of a theory is that, as models, they offer more opportunities for testing various more specific mechanisms or hypotheses. For example, what kinds of language input are most critical for developing the ability to reason about mental states? Past work has argued for the importance of at least three distinct sources, including exposure to mental state verbs \citep{brown_1996_WhyTalkMental}, the structure of interactive conversation \citep{harris_2005_ConversationPretenseTheory}, and certain syntactic constructions \citep{hale_2003_InfluenceLanguageTheory}.  Future work could compare different models with different training corpora (e.g., primarily dialogue vs. essays) to help isolate how much information is provided by each source of linguistic experience.

While these objections highlight the importance of future theoretical and empirical work, we believe that evidence for the \textit{sufficiency} of distributional information for competent False Belief task performance is a critical step toward assessing the plausibility of experience-based theories of belief attribution in humans.

\section{Conclusion}

Where does the human ability to reason about beliefs of others come from? It could emerge in part from an innate, biologically evolved capacity \citep{bedny_2009_GrowingBlindDoes}. It might also depend on experience, including language input \citep{devilliers_2014_RoleLanguageTheory}. The current results help quantify the contribution of language input. On a text-based version of the False Belief Task, humans responded correctly (i.e., in a manner congruent with a character's belief states) $82.7\%$ of the time, while the largest LLM tested responded correctly $74.5\%$ of the time; additionally, LLM behavior did not fully explain human behavior. This suggests that language statistics alone are sufficient to generate \textit{some} sensitivity to false belief, but crucially, not to fully account for \textit{human} sensitivity to false belief. Thus, the ability of humans to attribute mental states to others may involve \textit{linking} this linguistic input to innate capacities or to other embodied or social experiences.

\bibliography{bibliography}

\end{document}



\maketitle


\section{Materials}

\subsection{False Belief Passages}

12 False Belief Task passage templates were constructed. Each contained two characters (A and B), two locations (Start and End), and an object (X). The characters were different genders to prevent pronoun gender ambiguity. In each passage, A places X in the Start location. End is also mentioned at the start of the passage, to ensure that both locations are mentioned an equal number of times. In order to manipulate the First Mention condition, we swapped the order of these two initial mentions of the location.

In False Belief passages, A then leaves the room (or other general area) that contains the Start location where X has been placed. B then moves the object from the Start to the End location. In True Belief passages, A leaves the room either before or after B moves the object, and A watches B move X from the Start to the End location. In order to manipulate the most Recent Mention of a location, we changed the order in which the objects were mentioned during the move (e.g. ``from A to B'', or ``to B from A''.)

The final sentence of each passage suggests that A believes X to be in a location. The location itself is omitted from the passage, so that participants could complete this sentence with whichever location they thought was most likely, and so that models could provide estimates of the probability of each location. We manipulated the explicitness of this Knowledge Cue. In the Explicit condition, the sentence states that A ``thinks'' X is in the omitted location. In the implicit condition, A ``goes to get'' X from the omitted location.

\subsection{Attention Check Questions}

For each passage, we designed two attention check questions to filter out data from bots or inattentive participants. Both questions asked participants about the ground truth state of the described world, and did not require any reasoning about mental states in order to answer correctly. The first question asked participants ``Where did A put the X at the beginning of the story?'' The second question asked ``Where was the X at the end of the story?''

\section{Procedure}

We recruited participants through Amazon's Mechanical Turk. Each participant was randomly assigned to one of 192 conditions, corresponding to one of the 192 passage versions (16 versions of 12 item templates). We recruited 6 participants for each condition, but Mechanical Turk slightly oversampled to 1161. All participants provided informed consent.


Participants first read the following instructions:

\begin{quote}
    In this experiment, you will first see a short story. Please read through the story once at your normal reading pace. Once you have read the story you will be asked to complete a sentence, which is a continuation of the story. Complete the sentence in the way that makes the most sense based on what you read in the story. Please use only a single word to complete the sentence. Finally you will be asked two questions about what happened in the story. Answer the questions with a single word.
\end{quote}

Participants were then presented with the passage (excluding the final sentence) and advanced to the questions when they had finished reading. They then saw the critical False Belief Task question: the final sentence of the passage with the location omitted. Participants were asked to complete the sentence with a single word using a free-response text box. Participants could not see the rest of the passage while answering the critical or attention check questions.

Participants then responded to the attention check questions on separate pages by typing their answers in free-response text boxes. They then completed a demographic survey that asked their age; gender; whether they had ever been diagnosed with dyslexia, autism spectrum disorder, or attention deficit disorder; whether they had normal or corrected-to-normal vision; and whether English was their native language. Finally participants completed a debrief that asked them what they thought the purpose of the experiment was.

\section{Analysis}

\subsection{GPT-3 Responses}

We presented each passage to GPT-3 \textit{text-davinci-002} (hereafter, GPT-3) using the OpenAI API. We presented each passage version twice, once with each of the possible locations (Start or End) as the final word in the passage. We took the log probability of each of the locations as the final token in the passage, and found the log-odds of the Start vs. End location: $\textrm{log}(\textrm{p}(\textrm{Start})) - \textrm{log}(\textrm{p}(\textrm{End}))$. This was also equal to the log-odds of Start vs. End in a restricted decision space where only the Start or End location was a possible completion. Where a location comprised multiple tokens, we summed the log probabilities of each token in the location word to find the total log probability for the location.

We then tested whether GPT-3 was sensitive to the Knowledge State manipulation (i.e. whether the log-odds of the character believing the object was in the Start location was significantly different in False Belief vs. True Belief passages).

To do this, we conducted a log-likelihood ratio test (LRT), which compared the likelihood of two linear mixed effects models using the lme4 package \citep{batesLme4Package2007}. Specifically, we constructed a full linear mixed effects model containing Knowledge State, along with the three other manipulated variables (First Mention, Recent Mention, and Knowledge Explicitness), as well as by-item random slopes for the effect of Knowledge State, and by-item random intercepts. We compared this model to a reduced model omitting only the effect of Knowledge State. Note that in the R code, the Knowledge State variable is called ``condition''.

The relevant R code is as follows:

\begin{verbatim}

model_all_fe = lmer(data = df,
                  log_odds ~ condition + knowledge_cue +
                    recent_mention + 
                    first_mention +
                    (1 + condition | item),
                  control=lmerControl(optimizer="bobyqa"),
                  REML = FALSE)


model_no_condition = lmer(data = df,
                  log_odds ~ knowledge_cue +
                    recent_mention + 
                    first_mention +
                    (1 + condition | item),
                  control=lmerControl(optimizer="bobyqa"),
                  REML = FALSE)

anova(model_all_fe, model_no_condition)

\end{verbatim}
 
Additionally, as a second-order question, we asked whether the effect of Knowledge State on log-odds varied as a function of whether the belief state was probed explicitly (e.g., “He thinks the object is in the \_\_\_”) or implicitly (e.g., “He goes to look for the object in the \_\_\_”).

Here, the relevant code was as follows:

\begin{verbatim}

model_full = lmer(data = df,
                  log_odds ~ condition * knowledge_cue +
                    recent_mention + 
                    first_mention +
                    (1 + condition | item),
                  control=lmerControl(optimizer="bobyqa"),
                  REML = FALSE)
                  
anova(model_full, model_all_fe)
                  
\end{verbatim}

Finally, we conducted subset analyses to detemine whether the effect of Knowledge State was present in each of the Explicit and Implicit Knowledge Cue conditions.

\subsection{Human Participant Responses}

We excluded 13 of 1156 participants because they said that they were non-native English speakers. We then preprocessed both critical and attention check responses by converting them to lowercase, removing the word ``the'' from the start of responses, and removing punctuation and whitespace. We then removed 513 participants who failed to answer both attention checks correctly. As the exclusion rate was unusually high, we manually coded incorrect attention check responses. 17\% contained more than a single word (violating instructions), 23\% contained nonsense responses (e.g. `good', `dfg', `peloponnesian'), 25\% were one of the start or end locations, but the incorrect one, and 35\% contained words that were related to the passage but not one of the start or end location. Although this was a large proportion of participants to exclude ($45\%$ of native-English-speaking participants), they seemed to have been genuinely inattentive participants. 75\% of incorrect attention check responses were neither the Start nor the End location, indicating that the participant had encoded very little about the passage. We perform exploratory analyses on these excluded responses below (§ Exploratory Analyses, Excluded Human Participants).

We implemented a parallel check for GPT-3, which responded correctly on both attention check questions for 86\% of the trials. Of the 26 Failed attention checks, 19 (10\%) of responses were mentions of one of the start or end locations, but the incorrect one, 5 (3\%) included more than one word, and 2 (1\%) contained words that were related to the passage but not one of the start or end location. 
We did not exclude these trials from our analysis as this would equate to removing entire items as opposed to specific participants who appeared to be inattentive. However, we report the results of an exploratory analysis on this subset of items below (§ Exploratory Analyses, Excluding Failed GPT-3 Attention Check Trials).

We further excluded 17 participants whose response to the critical location was neither the Start nor End location. This was done to ensure comparability to the GPT-3 results, where only the probability of the Start and End location were considered. We retained 613 participants after all exclusions.

In order to test whether human comprehenders were sensitive to Knowledge State, we used an LRT to compare the likelihood of binomial mixed effects models predicting participant responses with and without a fixed effect of Knowledge State. The base model had fixed effects of Knowledge Cue, Recent Mention, and First Mention; random intercepts by item; and random slopes for Knowledge State by item. The full model additionally had a fixed effect of Knowledge State:

\begin{verbatim}
model_all_but_lo = glmer(data = df,
                  response ~ condition + knowledge_cue+
                    recent_mention + 
                    first_mention +
                    (1 + condition| item),
                  control=glmerControl(optimizer="bobyqa"),
                  family = binomial())

model_all_but_lo_and_condition  = glmer(data = df,
                  response ~ knowledge_cue+
                    recent_mention + 
                    first_mention +
                    (1 + condition| item),
                  control=glmerControl(optimizer="bobyqa"),
                  family = binomial())


anova(model_all_but_lo, model_all_but_lo_and_condition)
\end{verbatim}

In order to test whether Knowledge State continued to explain variance once the GPT-3 log-odds of a response had been accounted for, we compared a base model with fixed effects of Knowledge Cue, GPT-3 log-odds, Recent Mention, and First Mention; random intercepts by item; and random slopes for Knowledge State by item, to a full model with an additional fixed effect of Knowledge State.

\begin{verbatim}
model_all_fe = glmer(data = df,
                  response ~ condition + knowledge_cue + log_odds +
                    recent_mention + 
                    first_mention +
                    (1 + condition| item),
                  control=glmerControl(optimizer="bobyqa"),
                  family = binomial())


model_no_condition = glmer(data = df,
                  response ~ knowledge_cue + log_odds +
                    recent_mention + 
                    first_mention +
                     (1 + condition| item),
                  control=glmerControl(optimizer="bobyqa"),
                  family = binomial())

anova(model_all_fe, model_no_condition)
\end{verbatim}

\section{Exploratory Analyses}

\subsection{Excluded Human Participants}

Due to the high exclusion rate introduced by our attention check questions, we performed additional exploratory analyses including these participants. We thank a reviewer for suggesting this follow-up analysis.

The mean accuracy of all native English-speaking participants, including all of those excluded by the attention check (n=1143), was 55.8\%. After excluding 270 responses that were neither of the start or end locations (n=873), accuracy was 73.1\%.

We reperformed our GPT-3 baseline analysis using all human responses (regardless of whether they passed the attention check) that were to either the start or end location (n=873). As in the original analysis there was a residual effect of Knowledge State (True vs. False Belief) in humans that could not be accounted for by GPT-3 predictions ($\chi^2(1) = 33.2, p < 0.001$).

\subsection{Excluding Failed GPT-3 Attention Check Trials}

We reperformed our analyses on the subset of 166  items (86\% of all items) for which GPT-3 answered both attention checks correctly, a parallel check to the one implemented for human participants. 
Mean accuracy for GPT-3 in these trials was 75.1\%.
As in the original analysis, GPT-3 Log-Odds of Start vs. End location was higher (i.e. Start was relatively more likely) in the False Belief than True Belief condition ($\chi^2(1) = 20.6, p < .001$), however, there was a residual effect of Knowledge State (True vs. False Belief) in humans that could not be accounted for by GPT-3 predictions ($\chi^2(1) = 41.1, p < 0.001$).

\section{Template Passages}

Example passages from each template. In all examples, the Knowledge State is False Belief, the First Mention is the Start location, the most Recent Mention is the End location, and the Knowledge Cue is Explicit.

\begin{enumerate}

\item Sean is reading a book. When he is done, he puts the book in the box and picks up a sweater from the basket. Then, Anna comes into the room. Sean leaves to get something to eat in the kitchen. While he is away, Anna moves the book from the box to the basket. Sean comes back into the room and wants to read more of his book. Sean thinks the book is in the \rule{0.5cm}{0.15mm}.

\item Mary is feeling hungry and decides to make a sandwich. She gets out the jam and puts some on her sandwich. When she has finished eating, she puts the jam away in the cupboard and gets a soda out of the fridge. Then, James walks into the kitchen. Mary goes out of the kitchen to get something from her room. While she is away, James moves the jam out of the cupboard and into the fridge. Mary comes back into the kitchen and decides to make another sandwich. Mary thinks the jam is in the \rule{0.5cm}{0.15mm}.

\item While Cameron is eating, he gets a stain on his shirt. He puts his shirt in the sink and picks up a sweater from the basket. Then he goes to his room to get changed. Cameron doesn't see Helen move the shirt from the sink to the basket. When Cameron comes back into the room he wants to wash the stain off his shirt. Cameron thinks the shirt is in the \rule{0.5cm}{0.15mm}.

\item Paula and Tim are playing catch with a football in the yard. After a while they get bored. Paula puts the football in the shed, and gets a bottle of water from the garage. Then Paula goes to run some errands. Paula is away while Tim moves the football out of the shed and into the garage. Paula gets back and wants to play catch again. Paula thinks the football is in the \rule{0.5cm}{0.15mm}.

\item Ed arrives home after a long day at work. He puts his keys in the hall and leaves his bag in the study. Seana arrives home a few minutes later. Afterwards, Ed goes to the bathroom. Ed doesn't see Seana move the keys from the hall to the study. When Ed gets back from the bathroom, he realizes he needs his keys. Ed thinks the keys are in the \rule{0.5cm}{0.15mm}.

\item Laura is in the study, using the stapler. When she is finished, she puts the stapler away in a drawer and the documents away in a cabinet. Then Ross wanders into the study. Laura leaves to go and make a coffee. Laura doesn't see Ross take the stapler out of the drawer and put it in the cabinet. Laura comes back into the study and remembers she has one more document to staple together. Laura thinks the stapler is in the \rule{0.5cm}{0.15mm}.

\item David and Marta go out to get some wine for the party. When they get home, David stores the wine in the garage and grabs a drink from the fridge. Then, David goes out to get some snacks. While David is gone, Marta decides the wine would be best cooled, so she moves the wine out of the garage and into the fridge. David returns home and wants to put out the wine. David thinks the wine is in the \rule{0.5cm}{0.15mm}. 

\item Sarah is an artist who has just finished a new painting. She wants to display the new painting, and so she moves the painting to the hall, then grabs a book from the bedroom. She goes outside to read in the garden. Meanwhile, James sees the painting and decides it would look better elsewhere. James moves the painting from the hall to the bedroom. When Sarah gets back from reading, she goes to have a look at her painting. Sarah thinks the painting is in the \rule{0.5cm}{0.15mm}.

\item Lisa is cooking dinner in the kitchen. She uses the grater and puts it away in the cupboard, then takes out the spatula from the drawer. She briefly leaves the kitchen, and while she is away, Robert moves the grater from the cupboard to the drawer. Lisa comes back and realizes she needs to use the grater again. Lisa thinks the grater is in the \rule{0.5cm}{0.15mm}.

\item John and Karen are working on a building site. John is using a shovel to spread cement. When he is done, he leaves the shovel in the toolbox, then grabs his phone from the van. John goes to the office to talk to the boss. While John is away, Karen takes the shovel out of the toolbox to tidy up some edges. When Karen is finished, she puts the shovel away in the van. Then, John returns from the office. He needs to use the shovel again. John thinks the shovel is in the \rule{0.5cm}{0.15mm}. 

\item Patrick and Nicole are sitting at the train station, waiting for their train. Patrick takes out the ticket to check it and puts it away in the suitcase, then grabs his camera from the backpack. Then he gets up to take photos of the trains. Patrick doesn't see Nicole take the ticket out of the suitcase and place it in the backpack. Patrick comes back to sit at the table with Nicole. He wants to check the ticket one more time. Patrick thinks the ticket is in the \rule{0.5cm}{0.15mm}.

\item Hannah has just come back from a ride and is putting her horse away. She puts the saddle in the stable, then puts her backpack in the hut. Then she goes inside to have a bath. Meanwhile, Martin takes the saddle from the stable and puts it in the hut. When Hannah finishes her bath, she wants to use the saddle again. Hannah thinks the saddle is in the \rule{0.5cm}{0.15mm}.

\end{enumerate}

\section{Passages Versions}

All 16 versions of item 1. Conditions are marked as follows:

\begin{itemize}

    \item Knowledge State (KS) = \{False Belief (FB), True Belief (TB)\}
    \item First Mention (FM) = \{Start Location (S), End Location (E)\}
    \item Recent Mention (RM) = \{Start Location (S), End Location (E)\}
    \item Knowledge Cue (KC) = \{Explicit (E), Imlicit (I)\}
    
\end{itemize}

\begin{enumerate}

\item \textbf{[KS=FB, FM=S, RM=E, KC=E]} Sean is reading a book. When he is done, he puts the book in the box and picks up a sweater from the basket. Then, Anna comes into the room. Sean leaves to get something to eat in the kitchen. While he is away, Anna moves the book from the box to the basket. Sean comes back into the room and wants to read more of his book. Sean thinks the book is in the \rule{0.5cm}{0.15mm}.

\item \textbf{[KS=FB, FM=S, RM=E, KC=I]} Sean is reading a book. When he is done, he puts the book in the box and picks up a sweater from the basket. Then, Anna comes into the room. Sean leaves to get something to eat in the kitchen. While he is away, Anna moves the book from the box to the basket. Sean comes back into the room and wants to read more of his book. Sean goes to get the book from the \rule{0.5cm}{0.15mm}.

\item \textbf{[KS=TB, FM=S, RM=E, KC=E]}Sean is reading a book. When he is done, he puts the book in the box and picks up a sweater from the basket. Then, Anna comes into the room. Sean watches Anna move the book from the box to the basket. Sean leaves to get something to eat in the kitchen. Sean comes back into the room and wants to read more of his book. Sean thinks the book is in the \rule{0.5cm}{0.15mm}.

\item \textbf{[KS=TB, FM=S, RM=E, KC=I]} Sean is reading a book. When he is done, he puts the book in the box and picks up a sweater from the basket. Then, Anna comes into the room. Sean watches Anna move the book from the box to the basket. Sean leaves to get something to eat in the kitchen. Sean comes back into the room and wants to read more of his book. Sean goes to get the book from the \rule{0.5cm}{0.15mm}.

\item \textbf{[KS=FB, FM=S, RM=S, KC=E]} Sean is reading a book. When he is done, he puts the book in the box and picks up a sweater from the basket. Then, Anna comes into the room. Sean leaves to get something to eat in the kitchen. While he is away, Anna moves the book to the basket from the box. Sean comes back into the room and wants to read more of his book. Sean thinks the book is in the \rule{0.5cm}{0.15mm}.

\item \textbf{[KS=FB, FM=S, RM=S, KC=I]} Sean is reading a book. When he is done, he puts the book in the box and picks up a sweater from the basket. Then, Anna comes into the room. Sean leaves to get something to eat in the kitchen. While he is away, Anna moves the book to the basket from the box. Sean comes back into the room and wants to read more of his book. Sean goes to get the book from the \rule{0.5cm}{0.15mm}.

\item \textbf{[KS=TB, FM=S, RM=S, KC=E]} Sean is reading a book. When he is done, he puts the book in the box and picks up a sweater from the basket. Then, Anna comes into the room. Sean watches Anna move the book to the basket from the box. Sean leaves to get something to eat in the kitchen. Sean comes back into the room and wants to read more of his book. Sean thinks the book is in the \rule{0.5cm}{0.15mm}

\item \textbf{[KS=TB, FM=S, RM=S, KC=I]} Sean is reading a book. When he is done, he puts the book in the box and picks up a sweater from the basket. Then, Anna comes into the room. Sean watches Anna move the book to the basket from the box. Sean leaves to get something to eat in the kitchen. Sean comes back into the room and wants to read more of his book. Sean goes to get the book from the \rule{0.5cm}{0.15mm}.

\item \textbf{[KS=FB, FM=E, RM=E, KC=E]} Sean is reading a book. When he is done, he picks up a sweater from the basket and puts the book in the box. Then, Anna comes into the room. Sean leaves to get something to eat in the kitchen. While he is away, Anna moves the book from the box to the basket. Sean comes back into the room and wants to read more of his book. Sean thinks the book is in the \rule{0.5cm}{0.15mm}.

\item \textbf{[KS=FB, FM=E, RM=E, KC=I]} Sean is reading a book. When he is done, he picks up a sweater from the basket and puts the book in the box. Then, Anna comes into the room. Sean leaves to get something to eat in the kitchen. While he is away, Anna moves the book from the box to the basket. Sean comes back into the room and wants to read more of his book. Sean goes to get the book from the \rule{0.5cm}{0.15mm}.

\item \textbf{[KS=TB, FM=E, RM=E, KC=E]}Sean is reading a book. When he is done, he picks up a sweater from the basket and puts the book in the box. Then, Anna comes into the room. Sean watches Anna move the book from the box to the basket. Sean leaves to get something to eat in the kitchen. Sean comes back into the room and wants to read more of his book. Sean thinks the book is in the \rule{0.5cm}{0.15mm}.

\item \textbf{[KS=TB, FM=E, RM=E, KC=I]} Sean is reading a book. When he is done, he picks up a sweater from the basket and puts the book in the box. Then, Anna comes into the room. Sean watches Anna move the book from the box to the basket. Sean leaves to get something to eat in the kitchen. Sean comes back into the room and wants to read more of his book. Sean goes to get the book from the \rule{0.5cm}{0.15mm}.

\item \textbf{[KS=FB, FM=E, RM=S, KC=E]} Sean is reading a book. When he is done, he picks up a sweater from the basket and puts the book in the box. Then, Anna comes into the room. Sean leaves to get something to eat in the kitchen. While he is away, Anna moves the book to the basket from the box. Sean comes back into the room and wants to read more of his book. Sean thinks the book is in the \rule{0.5cm}{0.15mm}.

\item \textbf{[KS=FB, FM=E, RM=S, KC=I]} Sean is reading a book. When he is done, he picks up a sweater from the basket and puts the book in the box. Then, Anna comes into the room. Sean leaves to get something to eat in the kitchen. While he is away, Anna moves the book to the basket from the box. Sean comes back into the room and wants to read more of his book. Sean goes to get the book from the \rule{0.5cm}{0.15mm}

\item \textbf{[KS=TB, FM=E, RM=S, KC=E]} Sean is reading a book. When he is done, he picks up a sweater from the basket and puts the book in the box. Then, Anna comes into the room. Sean watches Anna move the book to the basket from the box. Sean leaves to get something to eat in the kitchen. Sean comes back into the room and wants to read more of his book. Sean thinks the book is in the \rule{0.5cm}{0.15mm}

\item \textbf{[KS=TB, FM=E, RM=S, KC=I]} Sean is reading a book. When he is done, he picks up a sweater from the basket and puts the book in the box. Then, Anna comes into the room. Sean watches Anna move the book to the basket from the box. Sean leaves to get something to eat in the kitchen. Sean comes back into the room and wants to read more of his book. Sean goes to get the book from the \rule{0.5cm}{0.15mm}.

\end{enumerate}


\bibliography{bibliography}